\documentclass[article,3p]{elsarticle}

\usepackage{lineno,hyperref}
\usepackage[english]{babel}
\usepackage{flushend}
\usepackage{url}
\usepackage{amsmath,amssymb,amsfonts}
\usepackage{bm}
\usepackage{booktabs}
\usepackage[table]{xcolor}
\usepackage{color}
\usepackage{dcolumn}
\usepackage{natbib}
\usepackage{tabularx}
\usepackage{float}
\usepackage[colorinlistoftodos]{todonotes}
\usepackage{multirow}
\usepackage{aeguill} 
\usepackage{stackengine}
\usepackage{setspace}
\usepackage[retain-explicit-plus,binary-units=true]{siunitx}
\usepackage{tabulary}
\usepackage{mathtools}
\usepackage{blkarray}
\usepackage{threeparttable}
\usepackage{comment}
\usepackage{subfigure}
\usepackage{rotating}
\usepackage{tablefootnote}
\usepackage[noend]{algpseudocode}
\usepackage[ruled,vlined]{algorithm2e}
\usepackage{listings}
\modulolinenumbers[5]

\makeatletter
\def\ps@pprintTitle{%
  \let\@oddfoot\@empty
  \let\@evenfoot\@oddfoot
}
\makeatother

\journal{Simulation Modelling Practice and Theory}

\begin{document}

\begin{frontmatter}

\title{Simulation, Modelling and Classification of Wiki Contributors: Spotting The Good, The Bad, and The Ugly}

\author[ESPaddress]{Silvia García-Méndez} \corref{mycorrespondingauthor}
\ead{sgarcia@gti.uvigo.es}
\author[PTaddress]{F\'{a}tima Leal}
\ead{fatimal@upt.pt}
\author[PT2address,PT3address]{Benedita Malheiro}
\ead{mbm@isep.ipp.pt}
\author[ESPaddress]{Juan Carlos Burguillo-Rial}
\ead{J.C.Burguillo@uvigo.es}
\author[PT1address,PT4address,PT3address]{Bruno Veloso}
\ead{brunov@upt.pt}
\author[IRLaddress]{Adriana E. Chis}
\ead{adriana.chis@ncirl.ie}
\author[IRLaddress]{Horacio Gonz\'{a}lez--V\'{e}lez}
\ead{horacio@ncirl.ie}

\address[ESPaddress]{Information Technologies Group, atlanTTic, University of Vigo, School of Telecommunications Engineering, Campus Lagoas-Marcosende, Vigo, 36310, Galicia, Spain}
\address[PTaddress]{REMIT, Universidade Portucalense, Porto, Portugal}
\address[PT1address]{Universidade Portucalense, Porto, Portugal}
\address[PT2address]{ISEP/IPP, School of Engineering, Polytechnic Institute of Porto, Porto, Portugal}
\address[PT3address]{INESC TEC, Porto, Portugal}
\address[PT4address]{Faculty of Economics, University of Porto, Portugal}
\address[IRLaddress]{Cloud Competency Centre, School of Computing, National College of Ireland, Dublin, Ireland}

\cortext[mycorrespondingauthor]{Corresponding author: sgarcia@gti.uvigo.es}

\begin{abstract}
Data crowdsourcing is a data acquisition process where groups of voluntary contributors feed platforms with highly relevant data ranging from news, comments, and media to knowledge and classifications. It typically processes user-generated data streams to provide and refine popular services such as wikis, collaborative maps, e-commerce sites, and social networks. Nevertheless, this \textit{modus operandi} raises severe concerns regarding ill-intentioned data manipulation in adversarial environments. 
This paper presents a simulation, modelling, and classification approach to automatically identify human and non-human (bots) as well as benign and malign contributors by using data fabrication to balance classes within experimental data sets, data stream modelling to build and update contributor profiles and, finally, autonomic data stream classification. By employing WikiVoyage--a free worldwide wiki travel guide open to contribution from the general public--as a testbed, our approach proves to significantly boost the confidence and quality of the classifier by using a class-balanced data stream, comprising both real and synthetic data. Our empirical results show that the proposed method distinguishes between benign and malign bots as well as human contributors with a classification accuracy of up to \SI{92}{\percent}.
\end{abstract}


\vfill 
{The Version of Record is available online at: \url{https://doi.org/10.1016/j.simpat.2022.102616}.\par}

\end{frontmatter}

\section{Introduction}
Crowdsourcing platforms primarily rely on data produced and shared by an extensive network of contributors, known as ``The Crowd''. Given the ubiquitous Internet access and the increasing user-generated data sources, the number of crowdsourcing systems has dramatically increased in recent times, significantly modifying the online behaviour of users and businesses. People no longer rely on expert advice but on the experiences of other users to make decisions on purchases, leisure, news, learning, and information in general.

Such platforms integrate Artificial Intelligence (\textsc{ai}) techniques such as data mining to swiftly process the crowd feedback and provide tailored up-to-date services on the fly using data streams. However, the processing opacity and the voluntary nature of crowdsourced data raise algorithmic transparency and data reliability concerns. On the one hand, the opacity of \textsc{ai} algorithms affects the interpretability of the results and can only be tackled by developers at the algorithmic design stage. On the other hand, data manipulation performed on behalf of third-party interests has become omnipresent in crowdsourcing platforms with fake news in social networks, biased feedback in evaluation-based platforms, and undesired (spam) content in wiki pages. Such underground activities, which negatively impact the reliability of the results, are typically performed at large scale by ill-intentioned bots in adversarial contexts.

Commonly construed as self-activated software agents programmed to run continuously in distributed network environments, bots perform tasks and make decisions on behalf of their creators without direct human intervention. They typically sense, perceive, and adapt to the context they operate. According to Tsvetkova \textit{et al.} (2017)~\cite{Tsvetkova2017}, all four main types of bots--content extractors, action executors, content generators, and human emulators--include benign and malign sorts. While benign bots tend to perform repetitive tasks to improve quality of service and system conditions, malign bots are designed to degrade content and system conditions by conveying false or biased information, altering system parameters, and/or tampering with data sources. 

This work addresses the real-time profiling and classification of Wikivoyage contributors into human and non-human (bots) as well as benign and malign contributors. The proposed method relies on data simulation and modelling of wiki streams to obtain an automatic classification of contributors. While the data fabrication balances classes to improve the consistency of the results, data stream modelling builds contributor profiles on the fly.

Therefore, this paper furnishes a data stream classification to identify deceitful sources in real-time. Specifically, contributors are modelled using review-based information (\textit{e.g.}, number of reviews, frequency of reverts, links, and number of characters inserted and deleted) and ORES edit quality\footnote{Available at \url{www.mediawiki.org/wiki/ORES}, December 2021.}, an Application Programming Interface (\textsc{api}) designed for wiki platforms, which helps to automate vandalism detection and removal. The model is incrementally updated with each incoming review. The classifier uses the contributor profile, first, to differentiate humans from bot users. Then, we employ a stack-based Machine Learning (\textsc{ml}) model, \textit{i.e.}, a two-level stacking system, to detect both bot/human users and positive/negative contributions.

The experiments have been conducted using a real Wikivoyage data set containing \num{417} bots and \num{46952} humans. Despite the imbalanced class distribution, the final two-level stacking classifier presents results between \SI{80}{\percent} and \SI{95}{\percent} concerning the accuracy and \textit{F}-measure (both macro and micro results per class). 
To balance classes within the experimental data set, this work adopts a data fabrication scenario based on synthetic-generated data. The real and synthetic data have been combined, improving the system performance and proving the solution efficiency. In summary, our proposal allows us to understand the behaviour of both benevolent and malevolent bots to pre-empt malign contributions and, ultimately, to prevent attacks to wiki pages.

The rest of this paper is organised as follows.
Section~\ref{sec:2} includes relevant related work concerning contributor modelling and classification. Section~\ref{sec:3} introduces the proposed approach, describing the profiling, classification, and data fabrication model.
Section~\ref{sec:4} presents the empirical evaluation results of the developed classifiers, including the simulation of synthetic data to balance the classes within the experimental data set.
Finally, Section~\ref{sec:5} concludes and highlights new perspectives to detect ill-intentioned contributors.

\section{Related work}
\label{sec:2}

Data crowdsourcing has been explored in multiple domains, ranging from knowledge sharing, collaborative maps, social interaction to personalised recommendation~\cite{egger2016open}. For example, Wikipedia\footnote{Available at \url{www.wikipedia.org}, December 2021.} and Wikivoyage\footnote{Available at \url{www.wikivoyage.org}, December 2021.} offer useful knowledge, which is continuously updated and improved by an invisible army of volunteer contributors;
Waze\footnote{Available at \url{www.waze.com}, December 2021.} assists drivers based on information provided in real-time by the crowd concerning traffic, accidents, obstacles, and road conditions; 
Facebook\footnote{Available at \url{www.facebook.com}, December 2021.} provides a way for people to communicate and socialise based exclusively on peer input; and
TripAdvisor\footnote{Available at \url{www.tripadvisor.com}, December 2021.} makes personalised recommendations relying solely on opinions and classifications shared by the crowd in the form of ratings, textual reviews, or photos. 
In this continuously evolving data crowdsourcing world, detecting automated data manipulators (bots and spammers) and their intentions (benign or malign) is essential to ensure data reliability. Detection methods search for hyperactivity indicators such as the number of page views, bounce rates, fake conversions, abnormal traffic spikes, and atypical average session duration\footnote{Available at \url{www.netacea.com/glossary/detect-bot-traffic}, December 2021.}. The accurate identification and classification of bots and sockpuppets can arguably help to mitigate the effects of malign data manipulation.

\subsection{Classification of contributors}

Wiki contributors can be humans or bots, and their contributions may be benign or malign~\cite{Kumar2017}. Bot activities are not all flagged and can be mistaken as human contributions. For example, Steiner \textit{et al.} (2014)~\cite{Steiner2014} identify wiki bots in real-time by checking the bot flag and the name ``bot''. The authors found in Wikipedia a linear relationship between the number of bots and the number of edits. The most active bots display a classic hockey stick activity curve with a long tail composed of slower bots responsible for corrective repetitive tasks.

Malign activities or vandalism correspond, according to Wikimedia\footnote{Available at \url{www.wikimedia.org}, December 2021.}, to edits made with deliberate malign intent, which differ from inadvertently damaging edits. To address Wiki vandalism detection, Adler \textit{et al.} (2011)~\cite{Adler2011} propose a trust-based approach called WikiTrust.

Since 2015, Wikimedia has provided the Objective Revision Evaluation Service (\textsc{ores}\footnote{Available at \url{https://ores.wikimedia.org}, December 2021.}) to automate vandalism detection and removal. This \textsc{ml} system predicts the quality of edits (quality of edit probability score) as they are made, as well as the quality of article drafts (quality of article probability score\footnote{Available at \url{www.mediawiki.org/wiki/ORES}, December 2021.}). The article quality model bases its predictions on structural characteristics of the article, such as the number of sections, references or infobox presence, since these features seem to correlate strongly with good writing and tone. \textsc{ores} accepts single or batch-edit submissions.
More specifically, given an edit, \textsc{ores} predicts three probabilities: (\textit{i}) whether or not an edit causes damage; (\textit{ii}) if it was saved in good faith; and (\textit{iii}) if the edit will eventually be reverted. Regarding the quality of an article draft, \textsc{ores} returns the probability distribution of being in one of the predefined classes (spam, vandalism, attack, or \textsc{ok}).
These scores can also be used to help identify malign bots. Curiously, Tsvetkova \textit{et al.} (2017)~\cite{Tsvetkova2017} concluded that even benign Wikipedia bots, which support the encyclopedia in tasks such as correcting spelling, maintaining links, or undoing digital vandalism on pages, often undo each other's edits. Works such as the one presented by Yang \textit{et al.} (2017)~\cite{Yang2017} use the power of the crowd to detect malign users. Their solution encompasses a mechanism to encourage users to report collaboratively malign behaviours.

Indeed, malign behaviours in social networks have been explored by many authors in the literature~\cite{Subrahmanian:2016}, some of them following an \textsc{ml} approach \cite{Choi2016}. Notably, Yamak \textit{et al.} (2016)~\cite{Yamak2016} focus on the detection of multiple online accounts (sockpuppet) created for purposes of deception on the English Wikipedia. The authors select the number of user contributions by namespace; the frequency of revert after each contribution; the average of bytes added and removed from each revision; the average contribution length in the same article; and the interval between user registration and user first contribution. Finally, the best results were obtained with the Random Forest algorithm~\cite{Schonlau2020}. 

When it comes to the popular Twitter\footnote{Available at \url{www.twitter.com}, December 2021.} platform, it is far from safe from this type of attack. In fact, Velayutham \textit{et al.} (2017)~\cite{Velayutham:2017} seek to detect such bots in Twitter using an \textsc{ml}-based approach to distinguish them from human profiles. Consequently, they identified the features related to user profile and content analysis used to compute \textit{botScore}. A similar work by Efthimion \textit{et al.} (2018)~\cite{Efthimion2018} goes beyond identifying bot users through \textsc{ml} and focuses on determining their prevalence in the platform. Lately, the widely used Botometer diagnostic tool\footnote{Available at \url{https://botometer.osome.iu.edu}, December 2021.} has been under study. In this vein, Rauchfleisch \textit{et al.} (2020)~\cite{Rauchfleisch2020} analysed its diagnostic ability over time, concluding that its thresholds are prone to variance with the consequent increase in false positives and false negatives.

Kumar \textit{et al.} (2016)~\cite{Kumar2016} seek to automatically identify fake articles in Wikipedia. They exploit not only textual content but also temporal data and network features. The authors report that the same group speedily writes insightful conclusions such as hoax articles of relatively new accounts with highly dense or overlapping networks. Conversely, Green \& Spezzano (2017)~\cite{Green2017} address the problem of identifying spam users on Wikipedia as a binary classification task based on user editing behaviour. They use as features the 
edit size (average size of edits, standard deviation of edit sizes, variance significance); edit time (average time between edits, the standard deviation of time between edits); links in edit (unique link ratio, link ratio in edits); talk page edit ratio and username. Moreover, they combine these features with the \textsc{ores} scores to improve the accuracy of the classification. Finally, Heindorf \textit{et al.} (2016, 2019)~\cite{Heindorf2016,Heindorf2019} adopt an online learning approach to the task of wiki vandalism detection.

More closely related to our work, Sarabani \textit{et al.} (2017)~\cite{Sarabadan2017} use a Random Forest implementation to classify vandal contributions in Wikidata with \SI{89}{\percent} recall and reducing by up to \SI{98}{\percent} the contributions which need to be reviewed. This work moved from text-based feature engineering and paid special attention to contextual features and user profiling. Finally, consideration should also be given to hybrid approaches such as the one presented by Zheng \textit{et al.} (2019)~\cite{ZhengLei2019}. Mainly, they created a 9-level bot taxonomy based on their behaviour in the English Wikipedia using both unsupervised rules and labelled data.

Concerning map-based crowdsourcing applications, Sanchez \textit{et al.} (2018) \cite{Sanchez2018} present a method to detect malign behaviours in Waze. The authors explore the Sybil attack, which uses a coordinated group to provide false information. The user behaviour is modelled using graphs where the weight of edges is computed combining distance, time, speed and number of votes.

More sophisticated approaches are also present in the literature. Hall \textit{et al.} (2018)~\cite{Hall2018} adopt a gradient boosting classifier to detect previously unidentified bots in wikis based on implicit behavioural and other informal editing characteristics.
The adopted model has 15 activity pattern features and 14 revision comment-based features. The latter set of features proved to be highly effective at predicting bot edits.

Furthermore, Zheng \textit{et al.} (2019)~\cite{ZhengPanpan2019} called attention to the need for sufficient training data to detect malign contributors in collaborative communities like wikis. They solved this issue by exploiting a Long Short-Term Memory auto-encoder trained solely with benign users. Then, based on the representations learned, the system uses a Generative Adversarial Network as a discriminator to spot malign contributors.

More recent works, such as Nikesh \textit{et al.} (2020)~\cite{Nikesh2020}, address the problem of identifying Wikipedia articles with undisclosed paid content and their contributors. Once again, the authors rely on content-derived features, user, and edit history patterns to build an unsupervised \textsc{ml} system. The authors create a graph where the nodes represent articles together with their PageRank and the edges represent contributors who have edited both articles. This article network has been used to obtain the article local clustering coefficient and PageRank. Specifically, they collect article-based features (age of user account at article creation, presence of infobox, number of references, photos, article categories, and content length); article network-based features (article PageRank, and local clustering coefficient); user-based features (username-based features, average size of added text, average time difference, ten-byte long edit ratio, and percentage of edits on User or Talk pages). Once again, the best results have been obtained with a Random Forest classifier and the selected features combined with \textsc{ores} article-based features. Finally, it is also worth mentioning the broad survey on fake news detection by Zhang \& Ghorbani (2020)~\cite{ZHANG2020} where they analyse the negative impact of misinformation on social media and distinguish between human and non-human (bots and cyborgs) creators.

\subsection{Research contribution}

This article analyses wiki-based contributions adopting data stream simulation, modelling and classification. Data stream modelling builds and continuously updates contributor profiles using the stream of contributions. The proposed profiling model employs the features provided by the surveyed works, \textit{i.e.}, review-based information (\textit{e.g.}, number of reviews, reverts, links, number of characters inserted or deleted, \textit{etc.}) and \textsc{ores} edit quality. 
Based on the statistical metrics of real data, our data fabrication produces a class-balanced data set that helps to improve the consistency of the results.
Finally, the data stream classification identifies both human and non-human contributors as well as benevolent and malevolent contributors.

Table~\ref{tab:Comparison} depicts a comparison of relevant related work that addresses contributor classification. The literature presents a significant number of approaches to classify contributors. However, they do not furnish the contribution type classification, \textit{i.e.}, detection of malevolent and benevolent contributors. Furthermore, the modelling of wiki streams for the online automatic contributor classification has not been thoroughly explored. To overcome this gap, this work addresses contributor profiling and classification to identify malign behaviours in wiki-based crowdsourcing platforms on the fly.

\begin{table}[!ht]
\centering
\caption{Comparison of contributor classification approaches.~\label{tab:Comparison}}
\small
\begin{tabular}{lcccc} 
\toprule
\multirow{2}{*}{\textbf{Approach}} & \textbf{Contributor} & \textbf{Contribution} & \textbf{Synthetic} & \textbf{Wiki} \\
& \textbf{Classification} & \textbf{Classification} & \textbf{Data} & \textbf{Streams} \\
\midrule
Adler \textit{et al.} (2011) \cite{Adler2011} & & \checkmark \\
\midrule

\textcolor{black}{Choi \textit{et al.} (2016)} \cite{Choi2016} & & \checkmark \\ \midrule

Efthimion \textit{et al.} (2018) \cite{Efthimion2018} & \checkmark \\
\midrule

Green \& Spezzano (2017) \cite{Green2017} & \checkmark \\ 
\midrule

Hall \textit{et al.} (2018) \cite{Hall2018} & \checkmark & \\
\midrule

Heindorf \textit{et al.} (2016, 2019) \cite{Heindorf2016,Heindorf2019} & & \checkmark & & \checkmark \\ 
\midrule

Kumar \textit{et al.} (2016) \cite{Kumar2016} & \checkmark \\ 
\midrule

Nikesh \textit{et al.} (2020) \cite{Nikesh2020} & \checkmark & \\
\midrule

\textcolor{black}{Rauchfleisch \& Kaiser (2020)} \cite{Rauchfleisch2020} & \checkmark \\ \midrule

\textcolor{black}{Sanchez \textit{et al.} (2018)} \cite{Sanchez2018} & & \checkmark \\ \midrule

Sarabani \textit{et al.} (2017) \cite{Sarabadan2017} & & \checkmark\\
\midrule

Steiner \textit{et al.} (2014) \cite{Steiner2014} & \checkmark & \\
\midrule

\textcolor{black}{Subrahmanian \textit{et al.} (2016)} \cite{Subrahmanian:2016} & \checkmark \\ \midrule

\textcolor{black}{Tsvetkova \textit{et al.} (2017)} \cite{Tsvetkova2017} & \checkmark & \checkmark \\ \midrule

Velayutham \textit{et al.} (2017) \cite{Velayutham:2017} & \checkmark\\
\midrule

Yamak \textit{et al.} (2016) \cite{Yamak2016} & \checkmark & & \\
\midrule

\textcolor{black}{Yang \textit{et al.} (2017)} \cite{Yang2017} & \checkmark & \checkmark \\ \midrule

\textcolor{black}{Zheng, Albano \textit{et al.} (2019)} \cite{ZhengLei2019} & \checkmark \\ \midrule

\textcolor{black}{Zheng, Yuan} \textit{et al.} (2019) \cite{ZhengPanpan2019} & \checkmark & \checkmark\\
\midrule

\rowcolor[gray]{.90}\textbf{Our proposal}	& {\large \checkmark} & {\large \checkmark} & 
{\large \checkmark} & {\large \checkmark} \\
\bottomrule
\end{tabular}
\end{table}

\section{Proposed method}
\label{sec:3}

\begin{figure*}[ht!]
\centering
\includegraphics[scale=0.14]{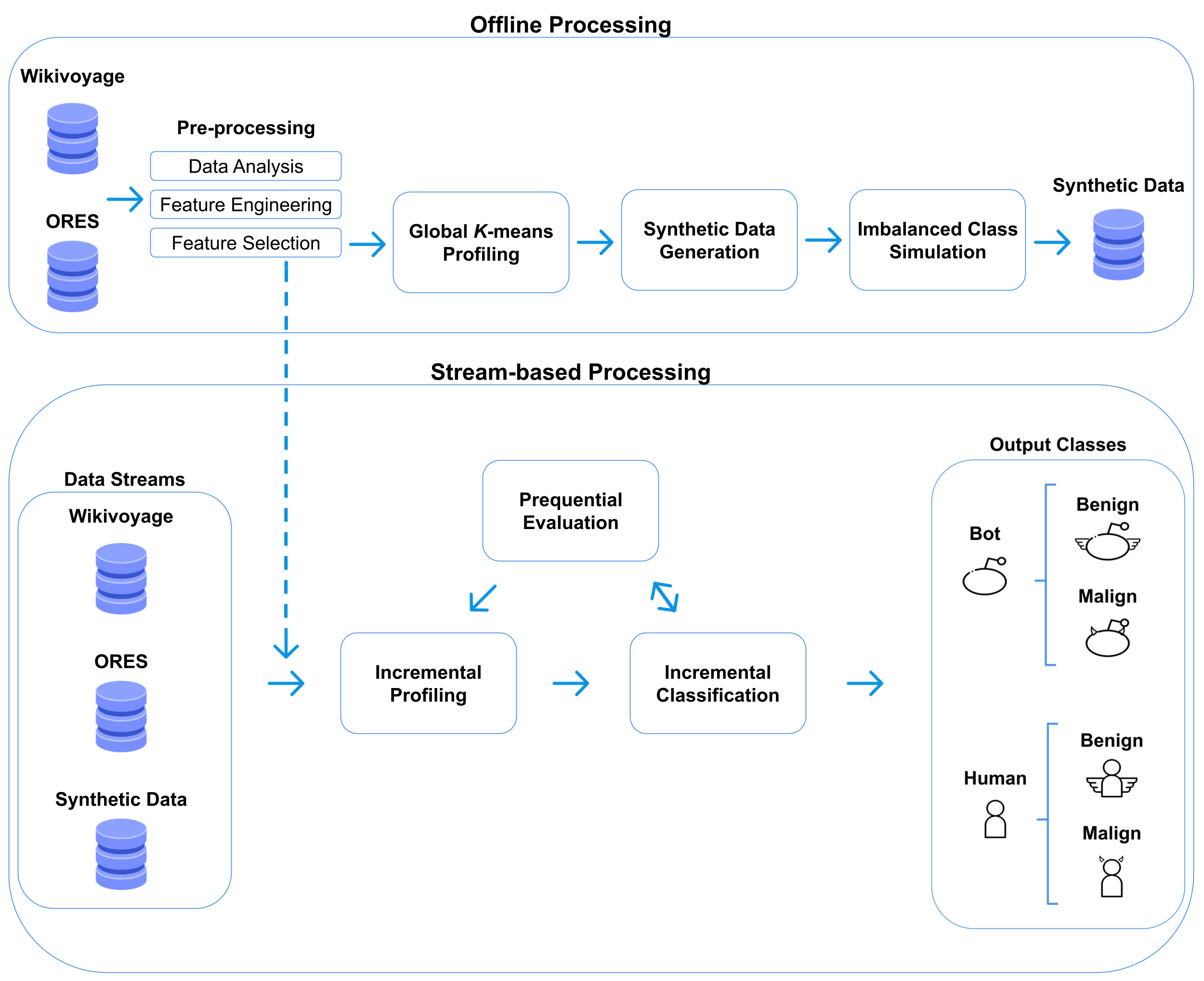}
\caption{\label{scheme}Proposed stream-based classification of wiki contributors.}
\end{figure*}

The proposed method is composed of multiple stages: (\textit{i}) pre-processing; (\textit{ii}) synthetic data generation; (\textit{iii}) classification; and (\textit{iv}) evaluation.
Figure~\ref{scheme} illustrates the solution designed to process wiki-based streams and generate classifications. The pre-processing stage (Section~\ref{sec:preprocessing}) analyses the input data in terms of pairwise correlations (Section~\ref{sec:data_analysis}) and employs feature engineering (Section~\ref{sec:feature_engineering}) to select the best features for the contributor profiling (Section~\ref{sec:feature_selection}). The synthetic-generated data enables to balance the classes and improve the model effectiveness (Section~\ref{sec:synthetic_data}). The resulting data set, composed of real and synthetic data, is then incrementally processed to model (Section~\ref{sec:profiling_stage}) and classify (Section~\ref{sec:classification_stage}) the contributors. The classification, which relies on well-known algorithms, explores single-class (binary) as well as a novel multi-class (stacking) classification of contributors and their editions. Finally, the outcomes are evaluated through standard classification metrics (Section~\ref{sec:evaluation_stage}).

\subsection{Pre-processing}
\label{sec:preprocessing}

Pre-processing aims to obtain a consistent feature space and facilitate compatibility with the target classification. This stage assembles three tasks: (\textit{i}) data analysis; (\textit{ii}) feature engineering; and (\textit{iii}) feature selection.

\subsubsection{Data analysis}
\label{sec:data_analysis}

The data analysis starts with a pairwise correlation of the features enumerated in Table~\ref{tab:features}, which are based on the literature review.
Equation~(\ref{pearson}) describes the Pearson Correlation Coefficient~\cite{Benesty2009} used to calculate the correlations where $x$ and $y$ represent two different features. The correlation coefficient ranges from -1 to 1. When the relation between two features is inverse, the metric is negative, otherwise it is positive. 

\begin{equation}
r_{xy} = \frac{\sum (x_i - \overline{x}) (y_i - \overline{y})}{\sqrt{\sum (x_i - \overline{x})^2} \sqrt{\sum (y_i - \overline{y})^2}}
\label{pearson}
\end{equation}

\subsubsection{Feature engineering}
\label{sec:feature_engineering}

The Wikivoyage data were compiled by the authors between January 14th and June 21st 2020. Wikivoyage holds crowdsourced information related to cities, attractions, monuments, hotels, restaurants, cultural details, \textit{etc}. In particular, the retrieved data set includes \num{3369} pages, \num{47369} contributors, and \num{319856} reviews.

Table~\ref{tab:features} contains the manufactured features of this experimental data set. The first target corresponds to feature \#2, where zero represents human and one represents non-human (bot) contributors. Furthermore, the second target feature, named \textit{contribution type}, is derived from feature \#17 (\textsc{ok} probability) provided by \textsc{ores}. When this probability is above \SI{50}{\percent}, the contribution is positive or favourable, otherwise, negative or hostile. In particular, zero corresponds to positive and one to negative edits.

\begin{table}[!ht]
\scriptsize
\centering
\caption{\label{tab:features} Features selected for the classification of contributors and their contributions.}
\begin{tabular}{cl} 
\toprule
\bf \# Num. & \bf Feature name \\\hline
1 & User ID \\
2 & Bot flag \\
3 & Number of reviews \\
4 & Average length of reviews \\
5 & Number of pages \\
6 & Average number of revisions per page \\
7 & Frequency of revisions per week \\
8 & Number of pages revised per week \\
9 & Number of reverts of the contributor reviews \\
10 & Revert frequency of the contributor reviews \\
11 & Links ratio of the reviews \\ 
12 & Repeated links ratio of the reviews \\ 
13 & Amount of inserted characters \\ 
14 & Amount of deleted characters \\
15 & ORES edit quality: true/false damaging \& good faith probability \\ 
16 & Average item quality from ORES: A/B/C/D/E probability \\
17 & Average article quality from ORES: OK/attack/spam/vandalism probability \\ 
18 & Average article quality from ORES (wp10): B/C/FA/GA/start/stub probability\\
\bottomrule
\end{tabular}
\end{table}

\begin{figure*}[ht!]
\centering
\includegraphics[scale=0.12]{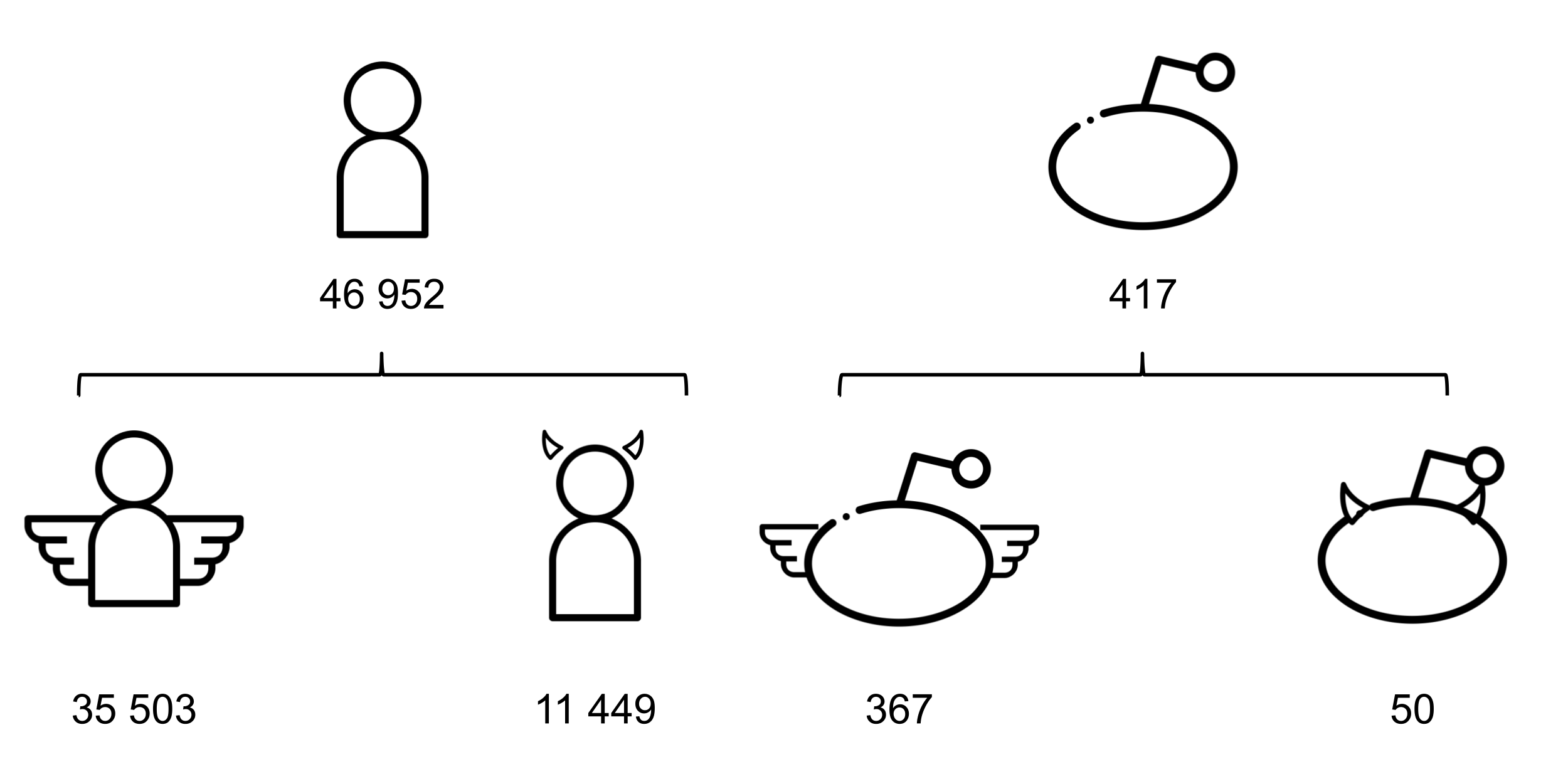}
\caption{\label{dataset}Distribution of the data set in terms of \textit{user type} and \textit{contribution type} targets.}
\end{figure*}

Figure~\ref{dataset} depicts the distribution of contributor classes within the original data set: (\textit{i}) \num{35503} benign and \num{11449} malign humans; and (\textit{ii}) \num{367} benign and \num{50} malign bots.

\subsubsection{Feature selection}
\label{sec:feature_selection}

Feature selection reduces the original feature space to improve performance and to avoid creating computationally unfeasible models. This work, due to the number of input features, adopted Recursive Feature Elimination (\textsc{rfe}\footnote{Available at \url{www.scikit-learn.org/stable/modules/generated/sklearn.feature_selection.RFE.html}, December 2021.}), a wrapper-type feature selection algorithm, to identify the features that maximise performance. \textsc{rfe} was configured to wrap the Linear Support Vector Classifier\footnote{Available at \url{www.scikit-learn.org/stable/modules/generated/sklearn.svm.LinearSVC.html}, December 2021.} and use as parameters \texttt{penalty=l1}, \texttt{dual=False}, \texttt{step=0.05} and \texttt{n\_jobs=-1}.

\subsection{Synthetic data generation}
\label{sec:synthetic_data}

\textcolor{black}{Synthetic data generation enables the simulation of stochastic and multi-spectral layouts to enhance \textsc{ml} model testing. This common practice found in the literature~\citep{Wan2017,Jain2020,Kurup2020,Mukherjee2021} supports the generation of relevant scenarios that, in most cases, are absent in the real data. Furthermore, it produces anonymous data, thereby maintaining the privacy of information. This process balances experimental data sets with synthetic samples which, in turn, are used to build models that react and operate correctly under the considered scenario.}

In classification, it is desirable to have data equally distributed among classes, \textit{i.e.}, ensure the data is free of any class bias. The main benefit of this scenario, where classifier models are built with balanced experimental data, is that they display higher classification accuracy.

In this work, synthetic data generation constitutes a user-controlled, accurate, cost-effective, efficient and scalable way to balance the classes within the experimental data set.

Synthetic samples were created using a statistical approach based on the quartile distribution of the most relevant features for both targets. \textcolor{black}{In particular, the proposed synthetic data generation method creates feasible editor daily incremental activity samples corresponding to the two target categories. The generated samples are composed of the features in Table~\ref{tab:features} and based on statistical measures (quartile distribution, median, minimum and maximum values) considering four intervals: (\textit{i}) from minimum to the first quartile (Q1); (\textit{ii}) from Q1 to median; (\textit{iii}) from median to third quartile (Q3); (\textit{iv}) from Q3 to the maximum.}

\begin{figure*}[ht!]
\centering
\includegraphics[scale=0.5]{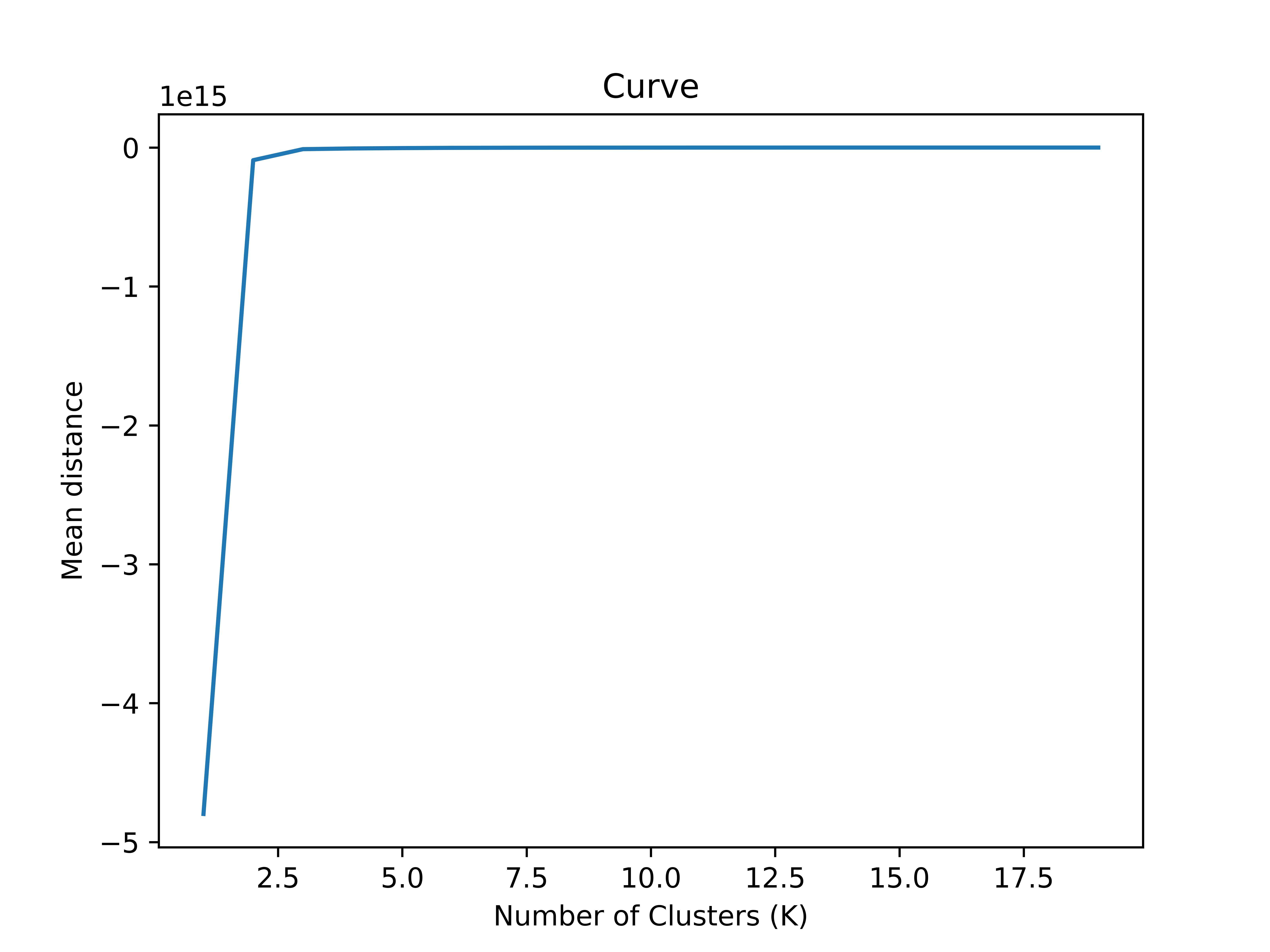}
\caption{\label{clustering}\textit{K}-means optimal \textit{K} value based on mean distance deviation.}
\end{figure*}

A non‐hierarchical \textit{K}‐means cluster analysis\footnote{Available at \url{www.scikit-learn.org/stable/modules/generated/sklearn.cluster.KMeans.html}, December 2021.} was performed on the original data set to identify common bot behaviour. Based on the information shown in Figure~\ref{clustering}, the model was configured to use two clusters ($K=2$) and trained with bot samples. 

\textit{K}-means provides the required statistic indicators to generate the new data, \textit{i.e.}, quartile distribution, minimum and maximum feature values. Algorithm~\ref{alg:building} describes the implemented synthetic data model. It generates \num{46532} new samples to balance the class distribution of the original data set. As previously mentioned, the model uses four different groups to create random feature values: (\textit{i}) from minimum to Quartile 1 (Q1); (\textit{ii}) from Q1 to median; (\textit{iii}) from median to Q3; (\textit{iv}) from Q3 to maximum.

\begin{algorithm*}[ht!]
 \small
 \caption{Synthetic data model}\label{alg:building}
 \DontPrintSemicolon
 \KwIn{\textit{K}-means statistic indicators}
 \KwData{$min$, $Q1$, $median$, $Q3$, $max$}
 \KwResult{Creates new samples for the experimental data set by generating feature values based on quartile distributions}
 \Begin{
 $count = \num{46532}$; \tcp*[h]{Number of samples to be generated}\; 
 
 $ranges = \{min, Q1, median, Q3, max\}$; \tcp*[h]{Quartile distribution}\; 
 
 $synthetic\_data = []$; \tcp*[h]{Synthetic samples}\; 
 
 \ForEach(){$r \in ranges-1$}{
 \ForEach(){$i \in count/4$}{
 $entry = random(ranges[r], ranges[r+1])$
 
 $synthetic\_data.append(entry)$
 }
 }
 }

 \tcp*[h]{Returns the synthetic generated data set} \; 
 \textbf{return} $synthetic\_data$ \;
\end{algorithm*}

Finally, the original and synthetic data are joined into the combined data set. The editions, characterised by the features in Table~\ref{tab:features}, are aggregated on a daily basis. The resulting data set is then processed as a stream.

\subsection{Incremental profiling}
\label{sec:profiling_stage}

The incremental profiling builds and continuously updates the profiles of contributors from the stream of samples stored in the combined data set. The profile of each contributor holds: (\textit{i}) the incremental sum of features \#3, \#5, \#9 and \#11-14; (\textit{ii}) the incremental average of features \#4, \#6, and \#15-18; and (\textit{iii}) the manufactured features \#7, \#8 and \#10. The latter are calculated as follows:

\begin{itemize}
 \item \#7 = \#3 / number of weeks, incremental sum.
 \item \#8 = \#5 / number of weeks, incremental sum.
 \item \#10 = \#9 / \#3, incremental sum.
\end{itemize}

\subsection{Classification}
\label{sec:classification_stage}

The classification task explores single-class (binary) and a novel multi-class (stacking) methods. Several stream-based binary classification algorithms were selected according to performance in similar problems~\cite{Heindorf2019,Salutari2020,Amaral2021} and availability in scikit-multiflow\footnote{Available at \url{https://scikit-multiflow.github.io}, December 2021.}, the adopted \textsc{ml} package for streaming data. They include single and ensemble methods:

\begin{itemize}
 \item Naive Bayes ({\sc nb})\footnote{Available at \url{https://scikit-multiflow.readthedocs.io/en/latest/api/generated/skmultiflow.bayes.NaiveBayes.html}, December 2021.}
 \item Decision Tree ({\sc dt})\footnote{Available at \url{https://scikit-multiflow.readthedocs.io/en/stable/api/generated/skmultiflow.trees.ExtremelyFastDecisionTreeClassifier.html}, December 2021.}
 \item Random Forest ({\sc rf})\footnote{Available at \url{https://scikit-multiflow.readthedocs.io/en/stable/api/generated/skmultiflow.meta.AdaptiveRandomForestClassifier.html\#skmultiflow.meta.AdaptiveRandomForestClassifier}, December 2021.} 
 \item Boosting Classifier ({\sc bc})\footnote{Available at \url{https://scikit-multiflow.readthedocs.io/en/stable/api/generated/skmultiflow.meta.OnlineBoostingClassifier.html}, December 2021.} 
\end{itemize}

\begin{figure*}[ht!]
\centering
\includegraphics[scale=0.13]{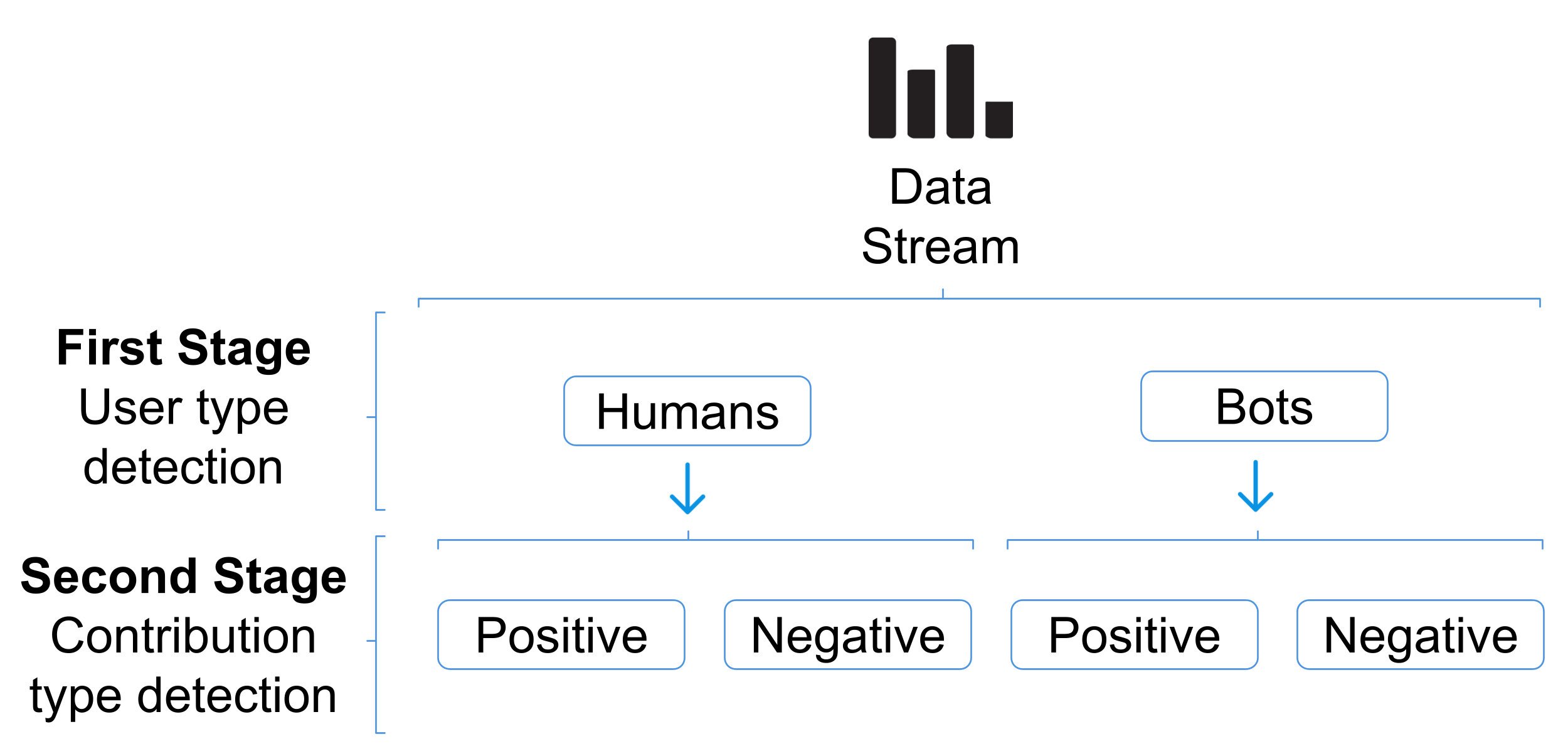}
\caption{\label{fig:stacking}Proposed data stream multi-class stacked classification of contributors.}
\end{figure*}

The proposed two-level stacking method, depicted in Figure~\ref{fig:stacking}, employs three \textsc{rf} models with the following hyperparameter configuration: \texttt{random\_state=0}, \texttt{n\_estimators=15}, \texttt{max\_features=None}.

\subsection{Evaluation metrics}
\label{sec:evaluation_stage}

The evaluation of the single-class (binary) and multi-class (stacking) classifiers uses classification accuracy, \textit{F}-measure, macro-averaging and micro-averaging.
The classification accuracy determines the classifier performance by measuring the number of correct classifications. 
The \textit{F}-measure combines Precision and Recall to establish the effectiveness of the classifier.
The micro-averaging and macro-averaging methods return a single value for the different metrics across multiple classes. While the macro-average computes the metric average independently for each class, \textit{i.e.}, treats all classes equally; the micro-average aggregates the contributions of all classes to compute the metric average. As suggested in the literature~\cite{TaoLiu2004,Liu2009}, the averaging methods should be used when the experimental data set is class-imbalanced.

\section{Experimental results}
\label{sec:4}

All experiments were performed on a server with the following hardware specifications:
\begin{itemize}
 \item Operating System: Ubuntu 18.04.2 LTS 64 bits
 \item Processor: Intel\@Core i9-9900K \SI{3.60}{\giga\hertz}
 \item RAM: \SI{32}{\giga\byte} DDR4 
 \item Disk: \SI{500}{\giga\byte} (7200 rpm SATA) + \SI{256}{\giga\byte} SSD
\end{itemize}

The experiments encompass: (\textit{i}) offline feature analysis to select the most relevant features to the contributor profile; (\textit{ii}) offline synthetic data generation to balance the classes within the data set; \textcolor{black}{and (\textit{iii}) classification of human and bot contributors as well as benign and malign contributions, \textit{i.e.}, regarding the two target features. Moreover, the classification stage explores three scenarios: (\textit{a}) stream-based binary classification of human and bot contributors using imbalanced real data; (\textit{b}) stream-based binary classification of benign and malign contributors using imbalanced real data; and (\textit{c}) stream-based multi-class stacked classification of benign and malign humans and bots using imbalanced (real) and balanced (real and synthetic) data. While scenarios (\textit{a}) and (\textit{b}) set the classification baseline with the original deeply unbalanced experimental data for each of the two target features, scenario (\textit{c}) analyses the performance of the stacking classification strategy in the simultaneous detection of contributor and contribution types.}

\textcolor{black}{Furthermore, the data collection\footnote{Available from the corresponding author on reasonable request.} relied on the well-known MediaWiki\footnote{Available at \url{www.pypi.org/project/mediawiki-utilities}, May 2022.}, a set of Python utilities for extracting and processing the features in Table~\ref{tab:features}.}
\textcolor{black}{The stream-based {\sc ml} models were incrementally updated and evaluated with \texttt{EvaluatePrequential}\footnote{Available at \url{https://scikit-multiflow.readthedocs.io/en/stable/api/generated/skmultiflow.evaluation.EvaluatePrequential.html}, May 2022.}. For each observed sample, the prequential evaluation makes a prediction, tests and trains the model.}

\subsection{\label{tuning_training}Feature analysis}
\label{sec:features}

To ensure good classification performance, feature pairwise correlation analysis was followed by feature selection.

\begin{description}

\item \textbf{Data analysis} detects the relevant features for contributor profiling. Therefore, the correlation between features and targets has been determined through the Pearson correlation coefficient. Table~\ref{heatmaps} shows for each target -- (1) \textit{user type}, \textit{i.e.}, human or non-human; and (2) \textit{contribution type}, \textit{i.e.}, malevolent or benevolent -- the features with a correlation below -0.15 or above 0.15. Note that these results indicate the current problem can be addressed with \textsc{ml} techniques as there exist moderate correlations with both targets.

\begin{table}[!ht]
\scriptsize
\caption{\label{heatmaps}Most correlated features with \textit{user type} (1) and \textit{contribution type} (2) targets.}
\begin{tabular}{ccccccccccc}
\toprule
\bf Target & \multicolumn{10}{c}{\bf Correlated features and correlation value} \\\midrule
\multirow{2}{*}{1} & \multicolumn{2}{l}{\#5} & \multicolumn{2}{l}{\#7} & \multicolumn{2}{l}{\#8} & \multicolumn{2}{l}{\#13} & \multicolumn{2}{l}{\#14} \\ \cmidrule(lr){2-11}
 & \multicolumn{2}{l}{0.16} & \multicolumn{2}{l}{0.55} & \multicolumn{2}{l}{0.57} & \multicolumn{2}{l}{0.16} & \multicolumn{2}{l}{0.22} \\\midrule
\multirow{2}{*}{2} & \#3 & \#5 & \#9 & \#13 & \#14 & \#15\tablefootnote{Damaging false probability.} & \#15\tablefootnote{Damaging true probability.} & \#18\tablefootnote{B probability.} & \#18\tablefootnote{C probability.} & \#18\tablefootnote{FA probability.}\\ \cmidrule(lr){2-11}
 & -0.24 & -0.24 & -0.18 & -0.20 & -0.17 & -0.16 & 0.16 & -0.18 & -0.17 & -0.19\\\bottomrule
\end{tabular}
\end{table}

\item \textbf{Feature selection}, as explained in Section~\ref{sec:feature_selection}, applies a combinatorial search over configurable parameter ranges and uses the \textsc{rfe} feature selection algorithm and LinearSVC to determine the best features. At the end, for the first target, the selected features (Table~\ref{tab:features}) were: \#6 to \#10, \#12, \#15 (\textit{good faith}), \#16 (\textsc{e} probability) and \#18 (\textsc{b} and \textit{stub} probabilities). Furthermore, the second target relies on \#18 (excluding \textsc{ga} probability).

\end{description}

\subsection{Synthetic data generation}
\label{sec:syntheticdata}

The quality of the new data is established by comparing the statistical properties and class distribution of both original and generated data.

\begin{description}
\item \textbf{Analysis of the generated synthetic data} is shown in Table~\ref{tab:comparative_original_synthetic}, a statistical comparison between the original and generated data. \textcolor{black}{Specifically, it shows the statistical attributes of the real data and the relative percentage change of the synthetic versus the original samples considering mean, minimum values, the first, second and third quartiles.} The results cover all relevant features for both targets, excluding \#15 (\textit{good faith}) \textcolor{black}{in Table~\ref{tab:features}} because synthetic and the original data are statistically identical. \textcolor{black}{Since the variations are minimal for most features, the results indicate that the proposed synthetic data generation algorithm maintains the statistical attributes of the real data. The sole exception is the number of reverted contributions per contributor (\#9 in Table~\ref{tab:features}) since it does not represent a probabilistic value.}

\begin{sidewaystable}[ph!]
\centering
\scriptsize
\caption{Statistical comparison between the original and the synthetic data \textcolor{black}{using mean, minimum, the first, second and third quartiles}.}\label{tab:comparative_original_synthetic}
\begin{tabular}{lllllllllllll}\toprule
\multicolumn{13}{c}{\bf Statistics for the original bot user samples} \\\hline
\bf Feature & \bf \#6 & \bf \#7 & \bf \#8 & \bf \#9 & \bf \#10 & \bf \#12 & \bf \#16 (E) & \bf \#18 (B) & \bf \#18 (C) & \bf \#18 (FA) & \bf \#18 (Start) & \bf \#18 (Stub) \\\hline
mean & 1.55 & 195.53 & 137.84 & 5.03 & 0.01 & 0.07 & 0.94 & 0.20 & 0.06 & 0.01 & 0.31 & 0.34 \\
min & 1.00 & 1.00 & 1.00 & 0.00 & 0.00 & 0.00 & 0.75 & 0.01 & 0.01 & 0.00 & 0.02 & 0.01 \\
Q1 & 1.03 & 40.50 & 30.83 & 0.00 & 0.00 & 0.00 & 0.94 & 0.18 & 0.06 & 0.01 & 0.30 & 0.29 \\
Q2 & 1.13 & 77.50 & 51.75 & 2.00 & 0.00 & 0.00 & 0.95 & 0.21 & 0.06 & 0.01 & 0.31 & 0.32 \\
Q3 & 1.60 & 289.75 & 127.68 & 6.00 & 0.01 & 0.00 & 0.96 & 0.22 & 0.07 & 0.01 & 0.33 & 0.37 \\\hline

\multicolumn{13}{c}{\bf \% $\Delta$ Statistics for the bot user generated samples} \\\hline
mean & -13.60 & -13.36 & -41.42 & -37.90 & 1073.63 & -100.00 & -0.97 & 25.22 & 11.86 & 48.30 & -2.74 & 8.56\\
min & 0.00 & 0.88 & 0.66 & 0.00 & 0.00 & 0.00 & 0.01 & 0.42 & 0.14 & 0.00 & 0.31 & 0.91\\
Q1 & 0.00 & 0.00 & 0.00 & 1.50 & 0.00 & 0.00 & 0.00 & 0.00 & 0.00 & 0.00 & 0.00 & 0.00\\
Q2 & 0.00 & 0.01 & 0.01 & -25.00 & 0.00 & 0.00 & 0.00 & 0.00 & 0.00 & 0.00 & 0.00 & 0.00\\
Q3 & 0.00 & 0.00 & 0.00 & -12.50 & 0.30 & 0.00 & 0.00 & 0.00 & 0.00 & 0.00 & 0.00 & 0.00\\\bottomrule
\end{tabular}
\end{sidewaystable}

\item \textbf{Clustering} is performed using \textit{K}‐means. Figure~\ref{dataset_balanced} shows the distribution of classes for the experimental data set after adding synthetic information. Note that the user type is determined by the bot flag, feature \#2 in Table~\ref{tab:features}, while the type of contribution is based on the \textsc{ok} probability (feature \#17 in Table~\ref{tab:features}). In the latter case, when the \textsc{ok} probability is above \SI{50}{\percent}, the contribution is positive, otherwise it is negative.

\begin{figure*}[ht!]
\centering
\includegraphics[scale=0.12]{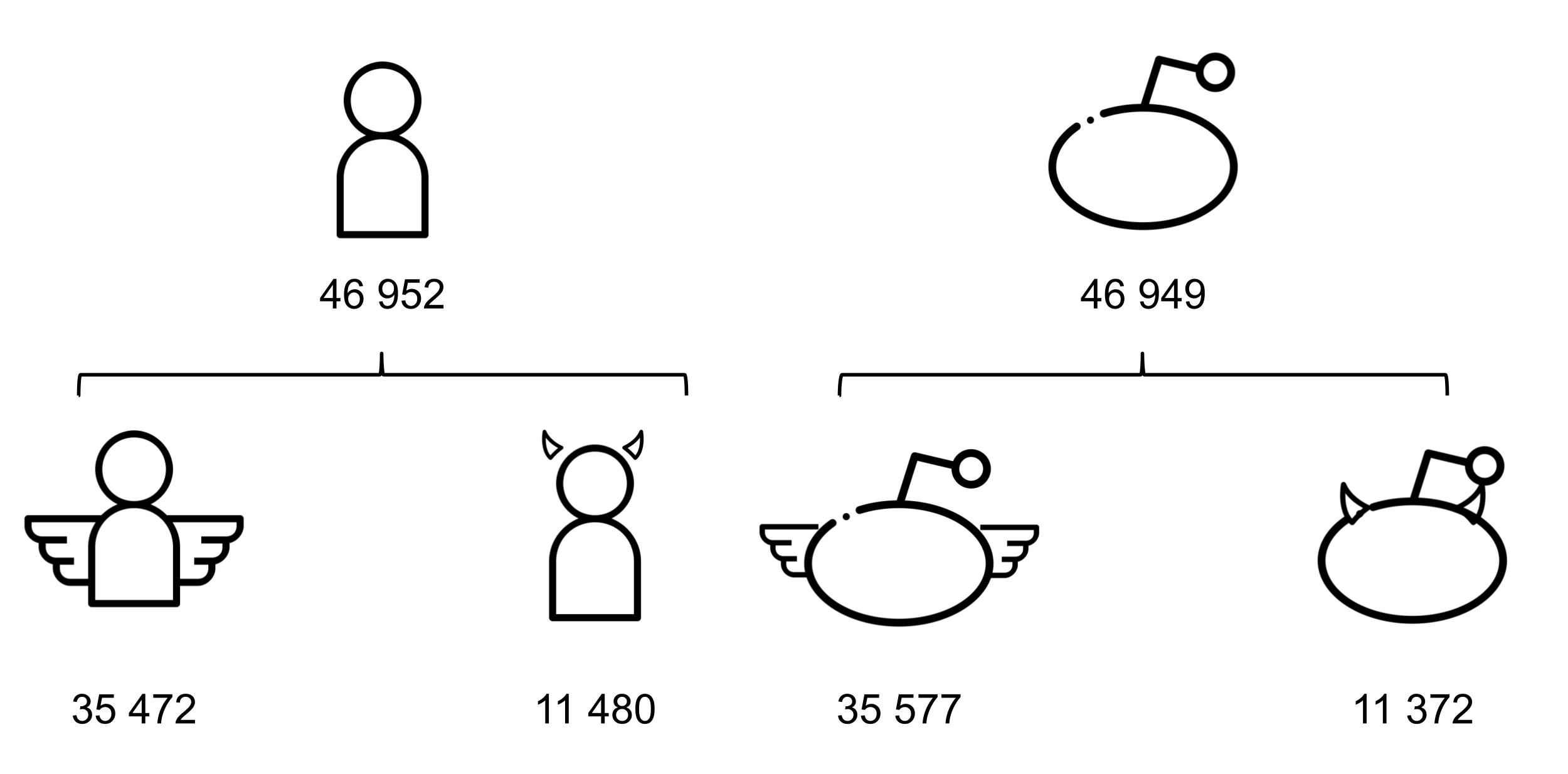}
\caption{\label{dataset_balanced}Distribution of \textit{user type} and \textit{contribution type} targets in the balanced data set.}
\end{figure*}

\end{description}

\subsection{Classification of human and bot contributors}
\label{sec:bots}

The proposed method estimates if the contributor is human or bot using the bot flag target feature. While class \#0 represents human contributors, class \#1 is bot contributors.
This experiment has been performed with a data stream produced from the original class-imbalanced data and three different sets of features:
\begin{enumerate}
 \item Basic set of features (features \#3 to \#14 from Table~\ref{tab:features});
 \item Full set of features (see Table~\ref{tab:features});
 \item Selected set of features (see Section~\ref{sec:features}). 
\end{enumerate}

Table~\ref{tab:scikitmultiflowResults_isbot} shows the macro and micro results of this classification. The values obtained are consistent in most cases for all classifiers and the values increase when using all features. Furthermore, the proposed method presents promising results when exclusively employing the most relevant features, since it reduces the processing time. The best binary classifier in all experiments is \textsc{rf}. Moreover, even though the original data set is deeply imbalanced, the system achieves a remarkable performance for both human and bot classes.

\begin{table}[!ht]
\centering
\scriptsize
\caption{\label{tab:scikitmultiflowResults_isbot}Performance of the human / bot classification.}
\begin{tabular}{ccccccS}
\toprule
\bf \multirow{2}{*}{Set} & \bf \multirow{2}{*}{Classifier} & \bf \multirow{2}{*}{Accuracy} & \multicolumn{3}{c}{\bf \textit{F}-measure} & \bf \multirow{2}{*}{{Time} (s)}\\
 & & & Macro & \#0 & \#1 & \\\hline
\multirow{5}{*}{1} & \textsc{nb} & 95.55 & 62.04 & 97.71 & 26.37
& 4.37 \\
& \textsc{dt} & 99.52 & 84.93 & 99.76 & 70.09
& 180.79 \\
& \textsc{rf} & \bf 99.83 & \bf 94.86 & \bf 99.91 & \bf 89.82
& 129.78 \\
& \textsc{bc} & \bf 99.83 & 94.84 & \bf 99.91 & 89.77
& 144.20 \\\hline

\multirow{5}{*}{2} & \textsc{nb} & 93.13 & 57.86 & 96.41 & 19.31
& 7.23 \\
& \textsc{dt} & 99.50 & 83.97 & 99.74 & 68.19
& 387.38 \\
& \textsc{rf} & \bf 99.83 & 94.60 & \bf 99.91 & 89.30
& 164.06 \\
& \textsc{bc} & \bf 99.83 & \bf 94.67 & \bf 99.91 & \bf 89.44
& 263.68 \\\hline

\multirow{5}{*}{3} & \textsc{nb} & 97.40 & 68.55 & 98.67 & 38.42
& 4.21 \\
& \textsc{dt} & 99.55 & 84.91 & 99.77 & 70.04
& 196.16 \\
& \textsc{rf} & \bf 99.82 & \bf 94.40 & \bf 99.91 & \bf 88.89
& 138.84 \\
& \textsc{bc} & \bf 99.82 & 94.34 & 99.90 & 88.77
& 110.26 \\\bottomrule
\end{tabular}
\end{table}

\subsection{Classification of benign and malign contributors}
\label{sec:contribution_type} 

The benign and malign classification experiments comprise: (\textit{i}) stream-based classification of the original class-imbalanced data using binary classifiers; and (\textit{ii}) stream-based classification of the original class-imbalanced and the new class-balanced data using a multi-class \textsc{rf} stacking classifier.

\subsubsection{Binary classification}
Table~\ref{tab:scikitmultiflowResults_contribution_type} provides the macro and micro results for the second target, \textit{i.e.}, benevolent and malevolent contributors classification, with single and ensemble binary classifiers and different feature sets. This experiment was performed with a data stream produced from the original class-imbalanced data set and the three different sets of features.

In light of the results, we can conclude that distinguishing between positive and negative contributions is not as straightforward as identifying humans and bots. Nonetheless, the proposed method achieves \SI{89.32}{\percent} accuracy and \SI{84.64}{\percent} macro \textit{F}-measure with the \textsc{rf} classifier for the best feature set analysed.

\begin{table}[!ht]
\centering
\scriptsize
\caption{\label{tab:scikitmultiflowResults_contribution_type}Performance of the benevolent / malevolent classification.}
\begin{tabular}{ccccccS}
\toprule
\bf \multirow{2}{*}{Set} & \bf \multirow{2}{*}{Classifier} & \bf \multirow{2}{*}{Accuracy} & \multicolumn{3}{c}{\bf \textit{F}-measure} & \bf \multirow{2}{*}{Time (s)}\\
 & & & Macro & \#0 & \#1 & \\\hline
\multirow{5}{*}{1} & \textsc{nb} & 57.50 & 54.84 & 65.79 & 43.90
& 4.24 \\
& \textsc{dt} & 80.10 & 69.17 & 87.53 & 50.82
& 344.72 \\
& \textsc{rf} & \bf 88.17 & \bf 82.94 & \bf 92.39 & \bf 73.49
& 196.26 \\
& \textsc{bc} & 78.63 & 72.50 & 85.49 & 59.50
& 127.39 \\\hline

\multirow{5}{*}{2} & \textsc{nb} & 69.74 & 62.42 & 79.00 & 45.85
& 6.92 \\
& \textsc{dt} & 80.66 & 70.17 & 87.85 & 52.50
& 725.77 \\
& \textsc{rf} & \bf 89.32 & \bf 84.64 & \bf 93.11 & \bf 76.17
& 270.83 \\
& \textsc{bc} & 80.90 & 74.59 & 87.25 & 61.93
& 211.41 \\\hline

\multirow{5}{*}{3} & \textsc{nb} & 75.03 & 55.68 & 84.96 & 26.41
& 2.93 \\
& \textsc{dt} & 78.95 & 66.16 & 86.96 & 45.35
& 203.73 \\
& \textsc{rf} & \bf 88.02 & \bf 82.57 & \bf 92.32 & \bf 72.83
& 180.11 \\
& \textsc{bc} & 79.89 & 68.85 & 87.39 & 50.31
& 101.09 \\\bottomrule
\end{tabular}
\end{table}

\subsubsection{Multi-class classification}
\label{sec:stacking}

The stacking model allows the simultaneous classification of contributor type (humans versus bots) and contribution type (positive versus negative).
This experiment was performed with two data streams generated from the original class-imbalanced and the new class-balanced sets.
The goal is to analyse the impact of the balanced data set, comprising original and synthetically generated data.

Table~\ref{tab:stacking} contains the results of: (\textit{i}) the baseline model obtained with feature set 3, the \textsc{rf} classifier and a stream produced from original class-imbalanced data (see Table~\ref{tab:scikitmultiflowResults_isbot} and Table~\ref{tab:scikitmultiflowResults_contribution_type}); (\textit{ii}) the stacking model with a stream produced from the original class-imbalanced data; and (\textit{iii}) the stacking model with a stream produced from the class-balanced data. When compared with the baseline, the stacking model improves the results in both accuracy and \textit{F}-measure. Furthermore, the results show the positive impact of the class-balanced data set. The difference in processing time is due to the fact that the imbalanced stream has \num{47369} events and the balanced stream has \num{93901} events. Actually, the process time per event is identical: \SI{21}{\milli\second\per{event}} for the imbalanced and \SI{20}{\milli\second\per{event}} for the balanced data stream.

\begin{table}[!ht]
\centering
\scriptsize
\caption{\label{tab:stacking}Performance of the multi-class \textsc{rf} stacking classifier versus baseline.}
\begin{tabular}{lccccc}
\toprule
{\bf Model} & {\bf Accuracy} & \multicolumn{3}{c}{\bf \textit{F}-measure} & {\bf Time (s)} \\
& & Macro & \#0 & \#1 \\\hline
Baseline\tablefootnote{The baseline was extracted from Table \ref{tab:scikitmultiflowResults_contribution_type}, set 3.} & 88.02 & 82.57 & 92.32 & 72.83 & 180.11 \\
Original data set & 88.71 & 83.85 & 92.71 & 74.98 & 977.87\\
Balanced data set & \bf 91.44 & \bf 87.46 & \bf 94.52 & \bf 80.39 & 1834.64\\\bottomrule
\end{tabular}
\end{table}

\subsection{\textcolor{black}{Literature comparison}}
\label{sec:literature_comparison}

\textcolor{black}{The results of the proposed method are herein compared with some of the works listed in Table~\ref{tab:Comparison}. Most of the related search adopts offline processing. Note that online models are built from scratch and incrementally updated and evaluated, whereas the offline models are trained and then tested using distinct data partitions. Zheng \textit{et al.} (2019)~\cite{ZhengPanpan2019} detect the type of contributor and contributions with a similar performance in both accuracy and macro \textit{F}-measure to the proposed method. However, social media data should be treated as streams since new revisions arrive continuously, resulting in instant classifier updates~\cite{Heindorf2016,Heindorf2019}. Thus, having comparable results endorses the proposed method.}

\textcolor{black}{Taking into account solely contribution type detection, Adler \textit{et al.} (2011)~\cite{Adler2011} report \SI{99}{\percent} accuracy with a rather low recall metric of \SI{30}{\percent}, whereas Choi \textit{et al.} (2016)~\cite{Choi2016} match the same accuracy value without additional metrics for fair comparison. Kumar \textit{et al.} (2016)~\cite{Kumar2016} identify hoaxers from non-hoaxers with \SI{63}{\percent} accuracy (\num{26} and \num{28} percent points lower in accuracy for contribution type detection than our proposal as shown in Table~\ref{tab:scikitmultiflowResults_contribution_type} and Table~\ref{tab:stacking}, respectively). Compared to Green \& Spezzano (2017)~\cite{Green2017} and Velayutham \textit{et al.} (2017)~\cite{Velayutham:2017} contributor type detectors, our solution attains \num{+17} and \num{+9} percent points in accuracy for bot detection (see Table~\ref{tab:scikitmultiflowResults_isbot} and Table~\ref{tab:stacking}, respectively). Finally, Zheng, Albano \textit{et al.} (2019)~\cite{ZhengLei2019} also report worse results for contributor classification: almost \num{-10} and \num{-3} percent points regarding macro \textit{F}-measure for bot detection according to Table~\ref{tab:scikitmultiflowResults_isbot} and Table~\ref{tab:stacking}, respectively. Since the obtained online results contemplate all data samples rather than just the test partition samples, the proposed stream-based method considerably improves the task of classifying contribution and contributor types. Note that some works were not included in this analysis, \textit{e.g.},~\cite{Tsvetkova2017,Yang2017,Subrahmanian:2016,Rauchfleisch2020,Heindorf2016,Heindorf2019,Sanchez2018}, since they do not provide directly comparable metrics.}

\section{Conclusions}
\label{sec:5}

Wiki-based repositories are built by a network of contributors, known as the crowd. The crowd voluntarily shares information related to people, places, entities, \textit{etc}. This voluntary nature of crowdsourcing platforms facilitates unethical behaviours from ill-intentioned human and bot contributors, enabling fraudulent data manipulation to satisfy third party interests. This scenario raises reliability concerns regarding the data quality of wiki-based platforms. To mitigate these issues, this work proposes a stream-based method to automatically classify contributors and their contributions. Specifically, the overall solution is composed of two distinct components: (\textit{i}) offline data pre-processing and data simulation; and (\textit{ii}) stream-based contributor profiling and classification. Pre-processing encompasses data analysis, feature engineering, and feature selection to identify the most promising edition-related features for profiling. The simulation is based on synthetic data generation to solve the deep class imbalance of the original data set. The stream-based profiling incrementally builds the profiles of contributors based on the selected edit features. The classification explores different stream-based binary classifiers with both class-imbalanced and class-balanced data streams to finally choose a stacking multi-class classifier that simultaneously identifies benign and malign humans and bots.

The proposed method was tested and evaluated with a real data set from Wikivoyage holding contributions from \num{417} bots and \num{46952} humans, using classification accuracy and \textit{F}-measure metrics. The experiments were conducted using the original Wikivoyage data together with synthetic data as streams, which incrementally update contributor profiles and classifier models. The final two-level stacking classifier presents results that vary between \SI{80}{\percent} and \SI{95}{\percent} concerning accuracy and \textit{F}-measure, proving the efficiency of the method.

To sum up, this paper describes a solution that can be used to anticipate and isolate malevolent content produced by the identified malign contributors. As future work, the plan is to generate stream-based synthetic data as well as explain classifications, employing Natural Language Processing techniques to generate statements automatically. 

\section*{Acknowledgements}
This work has been supported by: (\textit{i}) Xunta de Galicia grant ED481B-2021-118, Spain; (\textit{ii}) National Funds through the FCT – Fundação para a Ciência e a Tecnologia (Portuguese Foundation for Science and Technology) as part of project UIDB/50014/2020; (\textit{iii}) CHIST-ERA and the Irish Research Council as part of the ``Smart Pharmaceutical Manufacturing (SPuMoNI)'' research project [Apr/2019 -- Dec/2022]; and (\textit{iv}) University of Vigo/CISUG for open access charge.

\bibliography{mybibfile}

\end{document}